\documentclass[runningheads]{llncs}

\usepackage{graphicx}
\usepackage[pdfauthor={Marco Pegoraro, Merih Seran Uysal, David Benedikt Georgi, Wil M.P. van der Aalst},
pdftitle={Text-Aware Predictive Monitoring of Business Processes},
pdfkeywords={Predictive Monitoring, Process Mining, Natural Language Processing, LSTM Neural Networks}]{hyperref}
\usepackage{xcolor}
\usepackage{textcomp}
\usepackage{bbding}
\usepackage[font=scriptsize,labelfont=bf]{caption}
\usepackage{amsmath}
\usepackage{amssymb}
\usepackage{dsfont}
\usepackage{tikz}
\usetikzlibrary{calc, shapes, arrows, positioning, patterns, decorations.pathreplacing, backgrounds, petri, automata, matrix, spy}
\usepackage{pgfplots}
\pgfplotsset{compat=1.15}
\usepackage{tabularx}
\usepackage{lmodern}
\usepackage{siunitx}
\usepackage{booktabs}
\usepackage{etoolbox}
\usepackage{subcaption}
\usepackage{csquotes}

\hypersetup{
	colorlinks,
	linkcolor={red!50!black},
	citecolor={blue!60!black},
	urlcolor={blue!80!black}
}

\def\bag{{\cal B}}

\begin{document}

\title{Text-Aware Predictive Monitoring\\ of Business Processes\thanks{
		%In \emph{(venue)} (abbreviation), year. DOI: (DOI reference). \textcopyright publisher.
		We thank the Alexander von Humboldt (AvH) Stiftung for supporting our research interactions.
		Please do not print this document unless strictly necessary.
}}

\author{Marco Pegoraro~\Envelope\orcidID{0000-0002-8997-7517} \and Merih Seran Uysal\orcidID{0000-0003-1115-6601} \and David Benedikt Georgi \and Wil M.P. van der Aalst\orcidID{0000-0002-0955-6940}}

\authorrunning{Pegoraro et al.}

\institute{Process and Data Science chair, RWTH Aachen University, Aachen, Germany\\
	\email{\{pegoraro, uysal, wvdaalst\}@pads.rwth-aachen.de, david.georgi@rwth-aachen.de}\\
	\url{http://www.pads.rwth-aachen.de/}
}

\maketitle

\begin{abstract}
The real-time prediction of business processes using historical event data is an important capability of modern business process monitoring systems. Existing process prediction methods are able to also exploit the data perspective of recorded events, in addition to the control-flow perspective. However, while well-structured numerical or categor\-ical attributes are considered in many prediction techniques, almost no technique is able to utilize text documents written in natural language, which can hold information critical to the prediction task. In this paper, we illustrate the design, implementation, and evaluation of a novel \emph{text-aware process prediction model} based on \emph{Long Short-Term Memory} (LSTM) neural networks and natural language models. The proposed model can take categorical, numerical and textual attributes in event data into account to predict the activity and timestamp of the next event, the outcome, and the cycle time of a running process instance. Experiments show that the text-aware model is able to outperform state-of-the-art process prediction methods on simulated and real-world event logs containing textual data.

\end{abstract}

\section{Introduction}
In recent years, a progressive and rapid tendency to digital transformation has become apparent in most aspects of industrial production, provision of services, science, education, and leisure. This has, in turn, caused the widespread adoption of new technologies to support human activities. A significant number of these technologies specialize in the management of enterprise business processes.

The need of analysis and compliance in business processes, united to a larger and larger availability of historical event data have stimulated the birth and growth of the scientific discipline of \emph{process mining}. Process mining enables the discovery of process models from historical execution data, the measurement of compliance between data and a process model, and the enhancement of process models with additional information extracted from complete process cases.

Advancements in process mining and other branches of data science have also enabled the possibility of adopting \emph{prediction} techniques, algorithms that train a mathematical model from known data instances and are able to perform accurate estimates of various features of future instances. In the specific context of process mining, \emph{predictive monitoring} is the task of predicting features of \emph{partial process instances}, i.e., cases of the process still in execution, on the basis of recorded information regarding complete process instances. Examples of valuable information on partial process instances are the next activity in the process to be executed for the case, the time until the next activity, the completion time of the entire process instance, and the last activity in the case (outcome). If accurately estimated, these case features can guide process owners in making vital decisions, and improve operations within the organization that hosts the process; as a result, accurate predictive monitoring techniques are widely desirable and a precious asset for companies and organizations.

\begin{figure}[h]
	\centering
	\includegraphics[width=\textwidth, keepaspectratio]{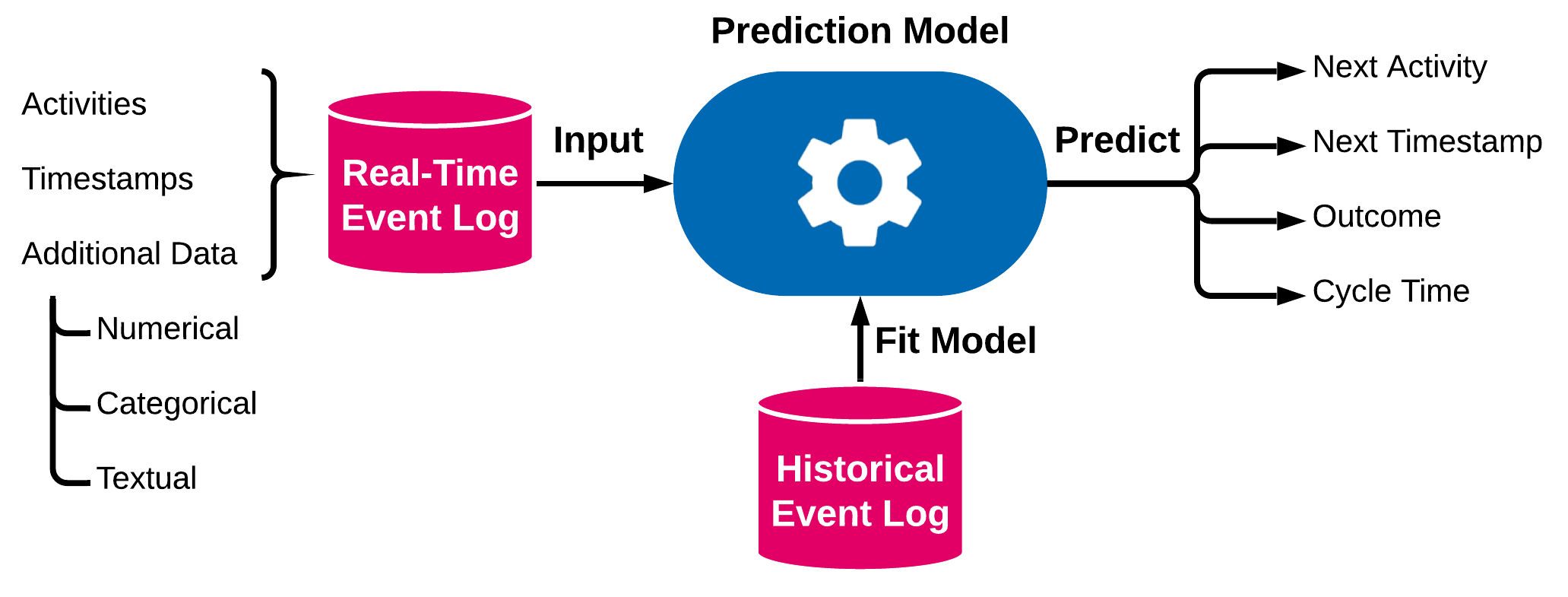}
	\caption{Problem overview: a general predictive monitoring model. The aim is predicting features of running process instances based on historical data, by exploiting numerical, categorical, and textual data.}
	\label{fig:problem-description}
\end{figure}

Existing predictive monitoring techniques typically operate at the merging point between process mining and machine learning, and are able to consider not only the control-flow perspective of event data (i.e., the activity, the case identifier, and the timestamp), but also additional data associated with them. However, few prediction techniques are able to exploit attributes in the form of text associated with events and cases. These textual attributes can hold crucial information regarding a case and its status within the workflow of a process. A general framework describing the problem is shown in Figure~\ref{fig:problem-description}.

The aim of this paper is to assess the extent to which textual information can influence predictive monitoring. To this end, we present a novel predictive monitoring approach able to exploit numerical, categorical, and textual attributes associated with events, as well as control-flow information. Our prediction model estimates features of cases in execution by combining a set of techniques for sequential and textual data encoding with predictions from an LSTM neural network, a machine learning technique particularly effective on sequential data such as process traces. Validation through experiments on real-life event logs shows that our approach is effective in extracting additional information from textual data, and outperforms state-of-the-art approaches for predictive monitoring.

The remainder of the paper is structured as follows. Section~\ref{sec:related} discusses some recent work related to predictive monitoring. Section~\ref{sec:prelim} presents some preliminary definitions. Section~\ref{sec:method} illustrates the details and architecture of our text-aware predictive monitoring technique. Section~\ref{sec:evaluation} presents the evaluation of the predictor and the results of the experiments. Section~\ref{sec:conclusion} concludes the paper.

\section{Related Work}\label{sec:related}
The intersection of process mining and machine learning is a rich and influential field of research. Among the numerous applications of machine learning in process mining, feature prediction on partial process traces based on historical complete traces (i.e., predictive monitoring) is particularly prominent.

Earlier techniques for prediction in process mining focused on white-box and human-interpretable models, largely drawn from statistics. Many proposals have been put forward to compute an estimate of the cycle time of a process instance, including decision trees~\cite{ceci2014completion} and simulation through stochastic Petri nets~\cite{rogge2013prediction}. Additionally, Teinemaa et al.~\cite{teinemaa2016predictive} proposed a process outcome prediction method based on random forests and logistic regression. Van der Aalst et al.~\cite{van2011time} exploit process discovery as a step of the prediction process, obtaining estimations through replay on an annotated transition system; this technique is then extended by Polato et al.~\cite{polato2018time} by annotating a discovered transition system with an ensemble of na\"ive Bayes and support vector regressors, allowing for the data-aware prediction of cycle time and next activity.

The second half of the 2010s saw a sharp turn from ensemble learning to single prediction models, and from white-box to black-box models -- specifically, recurrent neural networks. This is due to the fact that recurrent neural networks have been shown to be very accurate in learning from sequential data. However, they are not interpretable, and the training efficiency is often lower.

This family of prediction methods employs LSTM neural networks to estimate process instance features. Evermann et al.~\cite{evermann2016deep} proposed the use of LSTMs for next activity prediction; Tax et al.~\cite{tax2017predictive} trained LSTMs to predict cycle time of process instances. Navarin et al.~\cite{navarin2017lstm} extended this approach by feeding additional attributes in the LSTM, attaining data-aware prediction. More recently, Park and Song~\cite{park2020predicting} merged system-level information from a process model with a compact trace representation based on deep neural networks to attain performance prediction.

No existing predictive monitoring technique, to the best of our knowledge, incorporates information from free text, recorded as event or trace attribute, with the control-flow perspective of the process into a state-of-the-art LSTM neural network model for predictive monitoring: this motivates the approach we present in this paper.

\section{Preliminaries}\label{sec:prelim}
Let us first introduce some preliminary definitions and notations.

\begin{definition}[Sequence]
	A \emph{sequence} of length $n \in \mathbb{N}_0$ over a set $X$ is an ordered collection of elements defined by a function $\sigma \colon \{1, \dots, n\} \to X$, which assigns each index an element of $X$.
	A sequence  of length $n$ is represented explicitly as $\sigma = \langle x_1, x_2, \dots, x_n \rangle $ with $x_i \in X$ for $1 \leq i \leq n$. In addition, $\langle~\rangle$ is the empty sequence of length $0$. Over the sequence $\sigma$ we define $|\sigma| = n$, $\sigma(i) = x_i$, and $x \in \sigma \Leftrightarrow \exists_{1 \leq i \leq n} \colon x = x_i$.
	$X^*$ denotes the set of all sequences over $X$.
\end{definition}

The function $hd^k \colon X^* \to X^*$ gives the head or prefix of length $k$ of $\sigma$ for $0 \leq k \leq n$: $hd^k(\sigma)= \langle x_1, x_2, \dots, x_k\rangle$. For instance, $hd^2(\sigma)= \langle x_1, x_2 \rangle$.

\begin{definition}[Event, Trace, Event Log, Prefix Log]
	Let $\mathcal{A}$ be the universe of \emph{activity labels}. Let $\mathcal{T}$ be the closed under subtraction and totally ordered universe of \emph{timestamps}. Let $\mathcal{D}_1, \mathcal{D}_2, \dots , \mathcal{D}_m$ be the domains of \emph{additional attributes}. An \emph{event} is a tuple $e = (a, t, d_1, \dots, d_m) \in \mathcal{A} \times \mathcal{T} \times \mathcal{D}_1 \times \dots \times \mathcal{D}_m = \mathcal{E}$. Over an event $e$ we define the projection functions $\pi_\mathcal{A}(e) = a$, $\pi_\mathcal{T}(e) = t$, and $\pi_{\mathcal{D}_i}(e) = d_i$. A \emph{trace} $\sigma \in \mathcal{E}^*$ is a sequence of events such that time\-stamps are non-decreasing: $\pi_\mathcal{T} (e_i) \leq \pi_\mathcal{T} (e_j) $ for $1 \leq i < j \leq |\sigma|$. An \emph{event log} $L \in \bag(\mathcal{E}^*)$ is a multiset of traces. Given an event log $L$, we define the \emph{prefix log} $\mathbb{L} = \{ hd^k(\sigma) \mid  \sigma \in L \wedge 1 \leq k \leq |\sigma| \}$.
\end{definition}

Additional attributes $d_i \in \mathcal{D}_i$ may be in the form of text, i.e., its domain is the set of sequences $\mathcal{D}_i = \Sigma^*$ from a fixed and known alphabet $\Sigma$.

Next, let us define the target functions for our predictions:

\begin{definition}[Target Functions]
	Let $\sigma \in \mathcal{E}^*$ be a non-empty trace, and let $1 \leq k \leq |\sigma|$. The \emph{next activity} function $f_\mathrm{a}\colon \mathcal{E}^* \times \mathbb{N} \to \mathcal{A} \cup \{\blacksquare\}$ returns the activity of the next event, or an artificial activity $\blacksquare$ if the given trace is complete:
	\begin{align*}
	f_\mathrm{a}(\sigma, k) &= 
	\begin{cases}
	\blacksquare& \text{if $k = |\sigma|$} \\
	\pi_\mathcal{A}(\sigma(k+1)) & \text{else} 
	\end{cases}
	\end{align*}
	The \emph{next timestamp} function $f_\mathrm{t} \colon \mathcal{E}^* \times \mathbb{N} \to \mathcal{T}$ returns the time difference between the next event and last event in the prefix:
	\begin{align*}
	f_\mathrm{t}(\sigma, k) &=
	\begin{cases}
	0 & \text{if $k = |\sigma|$} \\
	\pi_\mathcal{T}(\sigma(k+1)) - \pi_\mathcal{T}(\sigma(k)) & \text{else} 
	\end{cases}
	\end{align*}
	The \emph{case outcome} function $f_\mathrm{o} \colon \mathcal{E}^* \to \mathcal{A}$ returns the last activity of the trace: $f_\mathrm{o}(\sigma) = \pi_\mathcal{A}(\sigma(|\sigma|))$. The \emph{cycle time} function $f_\mathrm{c} \colon \mathcal{E}^* \to \mathcal{T}$ returns the total duration of the case, i.e., the time difference between the first and the last event of the trace:
	$f_\mathrm{c}(\sigma) =  \pi_\mathcal{T}(\sigma(|\sigma|))-  \pi_\mathcal{T}(\sigma(1))$.
\end{definition}

The prediction techniques we show include the information contained in textual attributes of events. In order to be readable by a prediction model, the text needs to be processed by a \emph{text model}. Text models rely on a \emph{text corpus}, a collection of text fragments called \emph{documents}. Before computing the text model, the documents in the corpus are preprocessed with a number of normalization steps: conversion to lowercase, tokenization (separation in distinct terms), lemmatization (mapping words with similar meaning, such as ``diagnose'' and ``diagnosed'' into a single lemma), and stop word removal (deletion of uninformative parts of speech, such as articles and adverbs). These transformation steps are shown in Table~\ref{tab:text-preprocessing}.

\begin{table}[!htbp]
	\caption{Text preprocessing transformation of an example document containing a single sentence.}
	\begin{tabularx}{\textwidth}{l l p{8.5cm}}
		\toprule
		\textbf{Step} & \textbf{Transformation} & \textbf{Example Document}                                                       \\ \midrule
		0             & Original       &   \enquote{The patient has been diagnosed with high blood pressure.} \\
		1             & Lowercase               &   \enquote{the patient has been diagnosed with high blood pressure.} \\
		2 & Tokenization  & $\langle$\enquote{the}, \enquote{patient}, \enquote{has}, \enquote{been}, \enquote{diagnosed}, \enquote{with}, \enquote{high}, \enquote{blood}, \enquote{pressure}, \enquote{.}$\rangle$ \\
		3 & Lemmatization & $\langle$\enquote{the}, \enquote{patient}, \enquote{have},  \enquote{be}, \enquote{diagnose}, \enquote{with},  \enquote{high}, \enquote{blood}, \enquote{pressure}, \enquote{.}$\rangle$  \\
		4             & Stop word filtering     & $\langle$\enquote{patient}, \enquote{diagnose}, \enquote{high}, \enquote{blood}, \enquote{pressure}$\rangle$ \\ \bottomrule
	\end{tabularx}
	\label{tab:text-preprocessing}
\end{table}

In order to represent text in a structured way, we consider four different text models:

\emph{Bag of Words} (BoW)~\cite{brown1992class}: a model where, given a vocabulary $V$, we encode a document with a vector of length $|V|$ where the $i$-th component is the \emph{term frequency} (tf), the number of occurrences of the $i$-th term in the vocabulary, normalized with its \emph{inverse document frequency} (idf), the inverse of the number of documents that contain the term. This tf-idf score accounts for term specificity and rare terms in the corpus. This model disregards the order between words.

\emph{Bag of N-Grams} (BoNG)~\cite{brown1992class}: this model is a generalization of the BoW model. Instead of one term, the vocabulary consists of $n$-tuples of consecutive terms in the corpus. The unigram model ($n = 1$) is equivalent to the BoW model. For the bigram model ($n = 2$), the vocabulary consists of pairs of words that appear next to each other in the documents. The documents are encoded with the td-idf scores of their n-grams. This model is able to account for word order.

\emph{Paragraph Vector} (Doc2Vec)~\cite{le2014distributed}: in this model, a feedforward neural network is trained to predict one-hot encodings of words from their context, i.e., words that appear before or after the target word in the training documents. An additional vector, of a chosen size and unique for each document, is trained together with the word vectors. When the network converges, the additional vector carries information regarding the words in the corresponding document and their relationship, and is thus a fixed-length representation of the document.

\emph{Latent Dirichlet Allocation} (LDA)~\cite{blei2003latent}: a generative statistical text model, representing documents as a set of topics, which size is fixed and specified a priori. Topics are multinomial (i.e., categorical) probability distributions over all words in the vocabulary and are learned by the model in an unsupervised manner. The underlying assumption of the LDA model is that the text documents were created by a statistical process that first samples topic from a multinomial distribution associated with a document, then sample words from the sampled topics. Using the LDA model, a document is encoded as a vector by its topic distribution: each component indicates the probability that the corresponding topic was chosen to sample a word in the document. LDA does not account for word order.

In the next section, we will describe the use of text models in an architecture allowing to process a log to obtain a data- and text-aware prediction model.

\section{Prediction Model Architecture}\label{sec:method}
The goal of predictive monitoring is to estimate a target feature of a running process instance based on historical execution data. In order to do so, predictive monitoring algorithms examine \emph{partial traces}, which are the events related to a process case at a certain point throughout its execution. Obtaining partial traces for an event log is equivalent to computing the set of all prefixes for the traces in the log. Prefix logs will be the basis for training our predictive model.

In this paper, we specifically address the challenge of managing additional attributes that are textual in nature. In order to account for textual information, we need to define a construction method for fixed-length vectors that encode activity labels, timestamps, and numerical, categorical, and textual attributes.

Given an event $e = (a, t, d_1, \dots, d_m)$, its activity label $a$ is represented by a vector $\vec{a}$ using \emph{one-hot encoding}. Given the set of possible activity labels $\mathcal{A}$, an arbitrary but fixed ordering over $\mathcal{A}$ is introduced with a bijective index function $index_\mathcal{A} \colon \mathcal{A} \to \{1, \dots, |\mathcal{A}|\}$. Using this function, the activity is encoded as a vector of size $|\mathcal{A}|$, where the component $index_\mathcal{A}(\pi_\mathcal{A}(e))$ has value 1 and all the other components have value 0. The function $\mathds{1}_\mathcal{A}\colon \mathcal{A} \to \{0,1\}^\mathcal{A}$ is used to describe the realization of such one-hot encoding $\vec{a} = \mathds{1}_\mathcal{A}(\pi_\mathcal{A}(e))$ for the activity label of the event $e$.

In order to capture time-related correlations, a set of time-based features is utilized to encode the timestamp $t$ of the event. We compute a time vector $\vec{t} = (\hat{t}_1, \hat{t}_2, \hat{t}_3, \hat{t}_4, \hat{t}_5, \hat{t}_6)$ of min-max normalized time features, where $t_1$ is the time since the previous event, $t_2$ is the time since the first event of the case, $t_3$ is the time since the first event of the log, $t_4$ is the time since midnight, $t_5$ is the time since previous Monday, and $t_6$ is the time since the first of January. The min-max normalization is obtained through the formula
\begin{equation*}
\hat{x} = \dfrac{x-\min(x)}{\max(x) - \min(x)}
\end{equation*}
where $\min(x)$ is the lowest and $\max(x)$ is the highest value for the attribute $x$.

Every additional attribute $d_i$ of $e$ is encoded in a vector $\vec{d_i}$ as follows:
\begin{align*}
\vec{d_i} &=
\begin{cases}
\mathds{1}_{\mathcal{D}_i}(d_i) & \text{if $\mathcal{D}_i$ is categorical} \\
\hat{d}_i & \text{if $\mathcal{D}_i$ is numerical} \\
\textsc{TextModel}(d_1) & \text{if $\mathcal{D}_i$ is textual}
\end{cases}
\end{align*}

The encoding technique depends on the type of the attribute. Categorical attributes are one-hot encoded similarly to the activity label. Numerical attributes are min-max normalized: if the minimum and maximum are not bounded conceptually, the lowest or highest value of the attribute in the historical event log is used for scaling. Finally, if $\mathcal{D}_i$ is a textual model, it is encoded in a fixed-length vector with one of the four text models presented in Section~\ref{sec:prelim}; the documents in the text corpus for the text model consist of all instances of the textual attribute $\mathcal{D}_i$ contained in the historical log. This technique allows to build a complete fixed-length encoding for the event $e = (a, t, d_1, \dots, d_m)$, which we indicate with the tuple of vectors $\textit{enc}(e) = (\vec{a}, \vec{t}, \vec{d_1}, \dots, \vec{d_m})$.

This encoding procedure allows us to build a training set for the prediction of the target functions presented in Section~\ref{sec:prelim} utilizing an LSTM neural network.

\begin{figure}[!t]
	\centering
	\includegraphics[width=.8\textwidth, keepaspectratio]{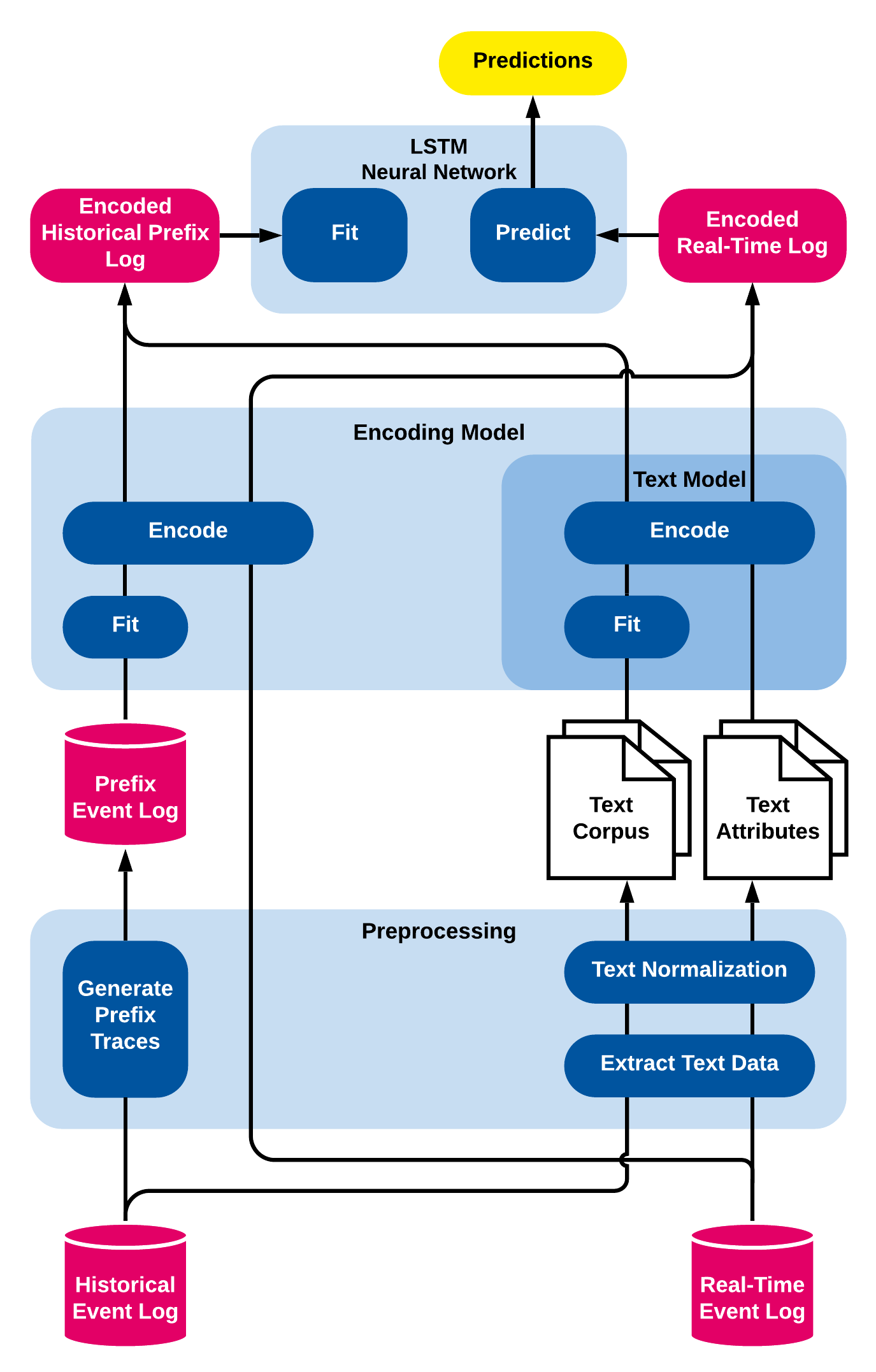}
	\caption{Overview of the text-aware process prediction model. Predictions for real-time processes are realized by an LSTM model that is fitted using an encoded representation of all prefixes of the historical event log. The encoding of textual attributes is realized by a text preprocessing pipeline and an exchangeable text encoding model.}
	\label{fig:framework}
\end{figure}

Figure~\ref{fig:framework} illustrates the entire encoding architecture, and the fit/predict pipeline for our final LSTM model. The schematic distinguishes between the \emph{offline} (fitting) phase, where we train the LSTM with encoded historical event data, and the \emph{online} (real-time prediction) phase, where we utilize the trained model to estimate the four target features on running process instances. Given an event log $L$, the structure of the training set is based on the partial traces in its prefix log $\mathbb{L} = \{ hd^k(\sigma) \mid \sigma \in L \wedge 1 \leq k \leq |\sigma| \}$. For each $\sigma = \langle e_1, e_2, \dots e_n \rangle \in L$ and $1 \leq k \leq n$, we build an instance of the LSTM training set. The network input $\langle vec{x_1}, \vec{x_2}, \dots, \vec{x_k} \rangle$ is given by the event encodings $vec{x_1} = \textit{enc}(e_1)$, $vec{x_2} = \textit{enc}(e_2)$, through $vec{x_k} = \textit{enc}(e_k)$. The targets $(\vec{y_a}, y_t, \vec{y_o}, y_c)$ are given by $\vec{y_a} = f_\mathrm{a}(\sigma, k)$, $\vec{y_t} = f_\mathrm{t}(\sigma, k)$, $\vec{y_o} = f_\mathrm{o}(\sigma)$, and $y_c = f_\mathrm{c}(\sigma, k)$.

\begin{figure}[h]
	\centering.
	\includegraphics[width=\textwidth, keepaspectratio]{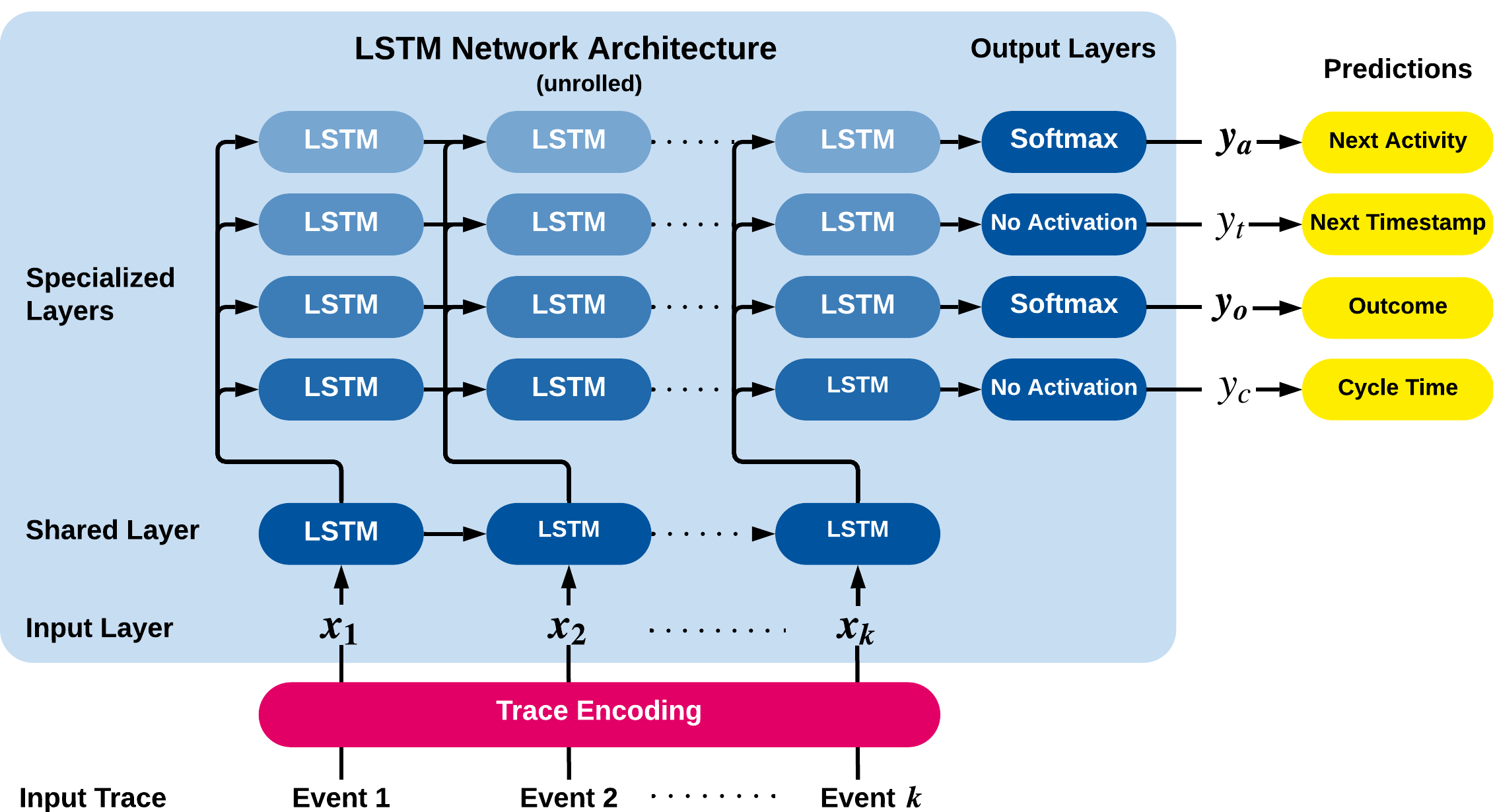}
	\caption{LSTM model architecture to simultaneously predict the next activity $(\vec{y_a})$, next event time $(y_t)$, outcome $(\vec{y_o})$ and cycle time $(y_c)$ for an encoded prefix trace $\vec{x_1}, \vec{x_2}, \dots, \vec{x_k}$.}
	\label{fig:network}
\end{figure}

Figure~\ref{fig:network} shows the topology of the network. The training utilizes gradient descent and backpropagation through time (BPTT). The loss for numerical prediction values $\hat{y}$ and the true value $y$ is the absolute error $\text{AE}(\hat{y},y)=|\hat{y} - y|$, while the loss for categorical prediction values is computed using the categorical cross-entropy error $\text{CE}(\vec{\hat{y}}, \vec{y}) = - \sum_{i=1}^{k} y_i \cdot \log \hat{y}_i$.

\section{Evaluation}\label{sec:evaluation}
The predictive monitoring approach presented in this paper has been implemented for validation, utilizing a Python-based, fully open-source technological stack. PM4Py~\cite{berti2019process} is a process mining Python tool developed by Fraunhofer FIT. It is used for event log parsing and its internal event log representation. The neural network framework Tensorflow~\cite{abadi2016tensorflow}, originally developed by Google, and its API Keras\footnote{\url{https://keras.io/}} were utilized to implement the final LSTM model. Furthermore, the libraries Scikit-learn~\cite{pedregosa2011scikit}, NLTK\footnote{\url{https://nltk.org/}}, and Gensim\footnote{\url{https://radimrehurek.com/gensim/}} provided the natural language processing capabilities required to preprocess and normalize text, as well as build and train the text models.

The text-aware model is compared to two other process prediction methods. First, the pure LSTM approach based on the ideas of Navarin et al.~\cite{navarin2017lstm} is considered, which only uses the activity, timestamp, and additional non-textual attributes of each event. This approach can be considered the state of the art in predictive monitoring with respect to prediction accuracy. The second baseline is the process model-based prediction method originally presented by van der Aalst et al.~\cite{van2011time}. This approach builds an annotated transition system for a log using a sequence, bag, or set abstraction. Each state of the transition system is annotated with measurements of historical traces that can be used to predict target values for unseen traces. During the prediction phase, running traces are mapped to the corresponding state of the transition system, and the measurements of the state are used to compute a prediction. We adopt the improvement of this method described in~\cite{tax2020interdisciplinary} to apply it to classification tasks and obtain the next activity and outcome predictions. The first 8 events of a trace are considered for the construction of the state space. Experiments with different horizon lengths (1, 2, 4, 16) mostly led to inferior results, and are thus not reported.

We evaluate the two baseline methods against our approach considering all four text models presented here, with a varying dimension of vector size (50, 100 and 500 for BoW and BoNG, 10, 20 and 100 for PV and LDA). The BoNG model is built with bigrams ($n = 2$). Of the four target functions presented in Section~\ref{sec:prelim}, classification tasks (next activity and outcome) are evaluated with a weighted-average class-wise F$_1$ score; regression tasks (next timestamp and cycle time) are evaluated on Mean Absolute Error (MAE). The first 2/3 of the chronologically ordered traces is used to fit the prediction model to the historical event data. The remaining 1/3 of traces are used to measure the prediction performance.

\begin{table}[t]
	\caption{Overview of the evaluated event logs with their key properties.}
	\setlength\tabcolsep{12pt}
	\begin{tabularx}{\textwidth}{l c c}
		\toprule
		\textbf{Event Log} & \textbf{Customer} & \textbf{Hospital}  \\
		& \textbf{Journey} &\textbf{Admission}  \\
		\midrule
		Cases & 15\,001& 46\,520\\
		Trace variants & 1001 &2784 \\
		Events & 55\,220 & 117\,952\\
		Events per case (mean) & 3.681& 2.536\\
		Median case duration (days) & 0.224& 7.579\\
		Mean case duration (days) &  0.713 & 121.154\\
		Activities & 18 & 26\\
		Words before preprocessing &247\,010 &  171\,938\\
		Words after preprocessing  &98\,915 & 165\,285\\
		Vocabulary before preprocessing & 1203 & 4973 \\
		Vocabulary after preprocessing & 817 & 4633\\
		Text attribute & Customer question & Diagnosis\\
		Additional non-textual attributes & Gender& Admission type\\
		& Age& Insurance\\
		\bottomrule
	\end{tabularx}
	\label{tab:logs}
\end{table}

\begin{table}[t]
	\tiny
	\setlength\tabcolsep{2pt}
	\caption{Snippet from the customer journey log.}
	\begin{tabularx}{\textwidth}{lllllp{4.7cm}}
		\toprule
		\textbf{Case} & \textbf{Activity}          & \textbf{Timestamp} & \textbf{Age} & \textbf{Gender} & \textbf{Message} \\
		\midrule
		40154127&question&2015/12/15 12:24:42.000&50-65&M&Can you send me a copy of the decision?\\
		40154127&taken&2015/12/30 15:39:36.000&50-65&M&\\
		40154127&mijn\_sollicitaties&2015/12/30 15:39:42.000&50-65&M&\\
		40154127&taken&2015/12/30 15:39:46.000&50-65&M&\\
		40154127&home&2015/12/30 15:39:51.000&50-65&M&\\
		23245109&question&2015/07/21 09:49:32.000&50-65&M&Law: How is the GAA (Average Number of Labor)?\\
		23245109&question&2015/07/21 09:54:28.000&50-65&M&Dismissal Procedure: Stops my contract automatically after two years of illness?\\
		23245109&question&2015/07/21 10:05:43.000&50-65&M&Dismissal: Am I entitled to a transitional allowance?\\
		23245109&question&2015/07/21 10:05:56.000&50-65&M&Chain Determination: How often may be extended a fixed-term contract?\\
		23245109&mijn\_werkmap&2015/07/27 09:54:03.000&50-65&M&\\
		23245109&mijn\_berichten&2015/07/27 09:54:13.000&50-65&M&\\
		23245109&mijn\_cv&2015/07/27 10:04:20.000&50-65&M&\\
		21537056&taken&2015/10/30 13:16:48.000&50-65&M&\\
		21537056&question&2015/10/30 13:22:00.000&50-65&M&How can I add a document/share with my consultant work through the workbook?\\
		21537056&taken&2015/10/30 13:23:24.000&50-65&M&\\
		21537056&mijn\_werkmap&2015/10/30 13:24:39.000&50-65&M&\\
		19290768&question&2015/09/21 12:41:21.000&30-39&V&Filling: What should I do if I made a mistake when filling out the Income Problem?\\
		19290768&home&2015/09/22 10:09:53.000&30-39&V&\\
		19290768&taken&2015/09/22 10:10:14.000&30-39&V&\\
		19290768&home&2015/09/22 10:11:12.000&30-39&V&\\
		53244594&mijn\_berichten&2016/02/25 09:10:40.000&40-49&M&\\
		53244594&question&2016/02/25 13:27:38.000&40-49&M&When is/are transferred my unemployment benefits?\\
		53244594&question&2016/02/29 10:04:23.000&40-49&M&Problem: I have to pay sv € 0 and further fill only the amount of holiday pay. What should I do if I get an error?\\
		53244594&question&2016/02/29 10:10:52.000&40-49&M&Why did you change the amount of my payment?\\
		\bottomrule
	\end{tabularx}
	\label{tab:snippet1}
\end{table}

\begin{table}[!h]
	\tiny
	\setlength\tabcolsep{2pt}
	\caption{Snippet from the hospital admission log.}
	\begin{tabularx}{\textwidth}{lllllp{1.8cm}}
		\toprule
		\textbf{Case} & \textbf{Activity}          & \textbf{Timestamp} & \textbf{Admission Type} & \textbf{Insurance} & \textbf{Diagnosis} \\
		\midrule
		8&PHYS REFERRAL/NORMAL DELI&2117-11-20 10:22:00&NEWBORN&Private&NEWBORN\\
		8&HOME&2117-11-24 14:20:00&NEWBORN&Private&\\
		9&EMERGENCY ROOM ADMIT&2149-11-09 13:06:00&EMERGENCY&Medicaid&HEMORRHAGIC CVA\\
		9&DEAD/EXPIRED&2149-11-14 10:15:00&EMERGENCY&Medicaid&\\
		10&PHYS REFERRAL/NORMAL DELI&2103-06-28 11:36:00&NEWBORN&Medicaid&NEWBORN\\
		10&SHORT TERM HOSPITAL&2103-07-06 12:10:00&NEWBORN&Medicaid&\\
		11&EMERGENCY ROOM ADMIT&2178-04-16 06:18:00&EMERGENCY&Private&BRAIN MASS\\
		11&HOME HEALTH CARE&2178-05-11 19:00:00&EMERGENCY&Private&\\
		12&PHYS REFERRAL/NORMAL DELI&2104-08-07 10:15:00&ELECTIVE&Medicare&PANCREATIC CANCER SDA\\
		12&DEAD/EXPIRED&2104-08-20 02:57:00&ELECTIVE&Medicare&\\
		13&TRANSFER FROM HOSP/EXTRAM&2167-01-08 18:43:00&EMERGENCY&Medicaid&CORONARY ARTERY DISEASE\\
		13&HOME HEALTH CARE&2167-01-15 15:15:00&EMERGENCY&Medicaid&\\
		16&PHYS REFERRAL/NORMAL DELI&2178-02-03 06:35:00&NEWBORN&Private&NEWBORN\\
		16&HOME&2178-02-05 10:51:00&NEWBORN&Private&\\
		17&PHYS REFERRAL/NORMAL DELI&2134-12-27 07:15:00&ELECTIVE&Private&PATIENT FORAMEN OVALE PATENT FORAMEN OVALE MINIMALLY INVASIVE SDA\\
		17&HOME HEALTH CARE&2134-12-31 16:05:00&ELECTIVE&Private&\\
		17&EMERGENCY ROOM ADMIT&2135-05-09 14:11:00&EMERGENCY&Private&PERICARDIAL EFFUSION\\
		17&HOME HEALTH CARE&2135-05-13 14:40:00&EMERGENCY&Private&\\
		18&PHYS REFERRAL/NORMAL DELI&2167-10-02 11:18:00&EMERGENCY&Private&HYPOGLYCEMIA SEIZURES\\
		18&HOME&2167-10-04 16:15:00&EMERGENCY&Private&\\
		19&EMERGENCY ROOM ADMIT&2108-08-05 16:25:00&EMERGENCY&Medicare&C 2 FRACTURE\\
		19&REHAB/DISTINCT PART HOSP&2108-08-11 11:29:00&EMERGENCY&Medicare&\\
		\bottomrule
	\end{tabularx}
	\label{tab:snippet2}
\end{table}

The process prediction models are evaluated on two real-world event logs, of which the general characteristics are given in Table~\ref{tab:logs}. Additionally, snippets of the datasets are shown in Tables~\ref{tab:snippet1} and~\ref{tab:snippet2}. The first describes the customer journeys of the Employee Insurance Agency commissioned by the Dutch Ministry of Social Affairs and Employment. The log is aggregated from two anonymized data sets provided in the BPI Challenge 2016, containing click data of customers logged in the official website werk.nl and phone call data from their call center.

The second log is generated from the MIMIC-III (Medical Information Mart for Intensive Care) database and contains hospital admission and discharge events of patients in the Beth Israel Deaconess Medical Center between 2001 and 2012.

The results of the experiments are shown in Table~\ref{tab:results}. The next activity prediction shows an improvement of 2.83\% and 4.09\% on the two logs, respectively, showing that text can carry information on the next task in the process. While the impact of our method on next timestamp prediction is negligible in the customer journey log, it lowers the absolute error by approximately 11 hours in the hospital admission log. The improvement shown in the outcome prediction is small but present: 1.52\% in the customer journey log and 2.11\% in the hospital admission log. Finally, the improvement in cycle time prediction is particularly notable in the hospital admission log, where the error decreases by 27.63 hours. In general, compared to the baseline approaches, the text-aware model can improve the predictions on both event logs with at least one parametrization.

\begin{table}[h]
	\caption{Experimental results for the next activity, next timestamp, outcome, and cycle time prediction. All MAE scores are in days.}
	\newrobustcmd{\B}{\fontseries{b}\selectfont}
	\setlength\tabcolsep{3.5pt}
	\scriptsize
	\begin{tabularx}{\textwidth}{
			>{\hsize=1.0\hsize}c
			>{\hsize=1.0\hsize}c|
			>{\hsize=1.0\hsize}c
			>{\hsize=1.0\hsize}c
			>{\hsize=1.0\hsize}c
			>{\hsize=1.0\hsize}c|
			>{\hsize=1.0\hsize}c
			>{\hsize=1.0\hsize}c
			>{\hsize=1.0\hsize}c
			>{\hsize=1.0\hsize}c
		}
		\toprule
		& & \multicolumn{4}{c|}{\textbf{BPIC2016 Customer Journey}} & \multicolumn{4}{c}{\textbf{MIMIC-III Hospital Admission}} \\
		Text & Text & Activity & Time & Outcome & Cycle & Activity & Time & Outcome & Cycle \\
		Model & Vect. Size & F$_1$ & MAE & F$_1$ & MAE & F$_1$ & MAE & F$_1$ & MAE \\
		\midrule
		\multicolumn{10}{c}{\textit{Text-Aware Process Prediction (LSTM + Text Model)}} \\
		BoW&50&     0.4251&     0.1764&     0.4732&     0.2357&     0.5389&    29.0819&     0.6120&    69.2953  \\
		BoW&100&     0.4304&  \B   0.1763&    0.4690&  \B   0.2337&     0.5487&    31.4378&   \B  0.6187&    70.9488 \\
		BoW&500&  \B   0.4312&     0.1798&     0.4690&     0.2354& \B    0.5596&    27.5495&     0.6050&    70.1084 \\
		BoNG&50&     0.4270&     0.1767&     0.4789&     0.2365&     0.5309&    27.5397&     0.6099&    69.4456 \\
		BoNG&100&     0.4237&     0.1770&     0.4819&     0.2373&     0.5450&    28.3293&     0.6094&    69.3619 \\
		BoNG&500&     0.4272&     0.1773&     0.4692&     0.2358&     0.5503&    27.9720&     0.6052&    70.6906 \\
		PV&10&     0.4112&     0.1812&     0.4670&     0.2424&     0.5265&    29.4610&     0.6007&    73.5219 \\
		PV&20&     0.4134&     0.1785&     0.4732&     0.2417&     0.5239&  \B  27.2902&     0.5962&    69.6191 \\
		PV&100&     0.4162&     0.1789&     0.4707&     0.2416&     0.5292&    28.2369&     0.6058&    69.4793 \\
		LDA&10&     0.4239&     0.1786&     0.4755&     0.2394&     0.5252&    28.8553&     0.6017&  \B  69.1465 \\
		LDA&20&     0.4168&     0.1767&     0.4747&     0.2375&     0.5348&    27.8830&     0.6071&    69.6269 \\
		LDA&100&     0.4264&     0.1777&   \B  0.4825&     0.2374&     0.5418&    27.5084&     0.6106&    69.3189 \\
		\multicolumn{10}{c}{\textit{LSTM Model Prediction Baseline}}  \\
		\multicolumn{2}{l|}{LSTM [9]} &     0.4029&     0.1781&     0.4673&     0.2455&     0.5187&    27.7571&     0.5976&    70.2978\\
		\multicolumn{10}{c}{\textit{Process Model Prediction Baseline (Annotated Transition System)}} \\
		\multicolumn{2}{l|}{Sequence [1, 13]}&     0.4005&     0.2387&     0.4669&     0.2799&     0.4657&    64.0161&     0.5479&   171.5684\\
		\multicolumn{2}{l|}{Bag [1, 13]} &     0.3634&     0.2389&     0.4394&     0.2797&     0.4681&    64.6567&     0.5451&   173.7963\\
		\multicolumn{2}{l|}{Set [1, 13]} &     0.3565&     0.2389&     0.4381&     0.2796&     0.4397&    63.2042&     0.5588&   171.4487\\
		\bottomrule
	\end{tabularx}
	\normalsize
	\label{tab:results}
\end{table}

\definecolor{bow}{RGB}{227,0,102}
\definecolor{bong}{RGB}{233,96,136}
\definecolor{pv}{RGB}{241,158,177}
\definecolor{lda}{RGB}{249,210,218}

\definecolor{lstm}{RGB}{100,100,100}

\definecolor{sequence}{RGB}{0,84,159}
\definecolor{bag}{RGB}{64,127,183}
\definecolor{set}{RGB}{142,186,229}

\pgfplotscreateplotcyclelist{colorlist}{%
	bow,line width=0.7pt,every mark/.append style={line width=160pt, fill=bow,},mark=o, mark options={scale=1.6}\\%1
	bong,line width=0.7pt,every mark/.append style={fill=bong},mark=diamond, mark options={scale=1.8}\\%2
	pv,line width=0.7pt,every mark/.append style={fill=pv},mark=otimes, mark options={scale=1.4}\\%3
	lda,line width=0.7pt,every mark/.append style={fill=lda},mark=star, mark options={scale=1.6}\\%4
	lstm,line width=0.7pt,every mark/.append style={fill=lstm},mark=square, mark options={scale=1.6}\\%5
	sequence,line width=0.7pt, every mark/.append style={fill=sequence},mark=10-pointed star, mark options={scale=1.2}\\%6
	bag,line width=0.7pt,every mark/.append style={fill=bag},mark=triangle, mark options={scale=1.8}\\%7
	set,line width=0.7pt,every mark/.append style={fill=set},mark=oplus, mark options={scale=1.3}\\%8
}

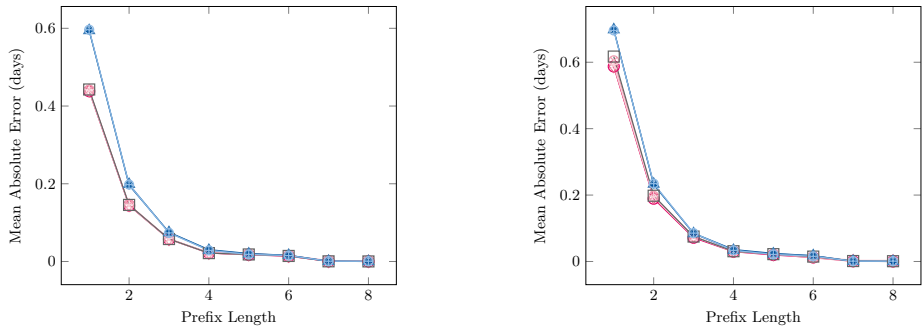
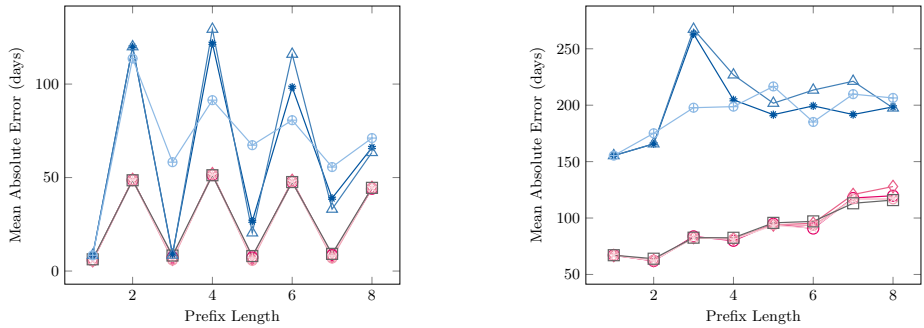
\begin{figure}[h]
	\centering
	\begin{subfigure}{\textwidth}
		\centering
		\begin{tikzpicture}[scale=.9]
		\begin{axis}[
		scale only axis,width=1mm, height=1mm,
		hide axis,
		draw=none,
		legend columns=-1,
		legend style={/tikz/every even column/.append style={column sep=0.2cm}, draw=none},
		cycle list name = colorlist,
		legend style={at={(10,5)}},
		legend image post style={sharp plot},
		]
		\pgfplotsinvokeforeach{1,...,8}{\addplot coordinates {(0,0)};} 
		\addlegendentry{BoW};
		\addlegendentry{BoNG};
		\addlegendentry{PV};
		\addlegendentry{LDA};
		\addlegendentry{LSTM};
		\addlegendentry{Sequence};
		\addlegendentry{Bag};
		\addlegendentry{Set};
		\end{axis}
		\end{tikzpicture}%
		\vfill
		\vspace{0.2cm}
		\begin{tikzpicture}[scale=0.65]
		\begin{axis}[
		xlabel={Prefix Length},
		ylabel={Mean Absolute Error (days)},
		cycle list name=colorlist,
		]
		\addplot table[x=index,y=ntBoW100, col sep= comma] {data/prefix_werk.csv};
		\addlegendentry{BoW}
		\addplot table[x=index,y=ntBoNG50, col sep= comma] {data/prefix_werk.csv};
		\addlegendentry{BoNG}
		\addplot table[x=index,y=ntPV20, col sep= comma]  {data/prefix_werk.csv};
		\addlegendentry{PV}
		\addplot table[x=index,y=ntLDA20, col sep= comma] {data/prefix_werk.csv};
		\addlegendentry{LDA}
		\addplot table[x=index,y=nt-0, col sep= comma] {data/prefix_werk.csv};
		\addlegendentry{LSTM}
		\addplot table[x=index,y=ntsequence8, col sep= comma] {data/prefix_werk.csv};
		\addlegendentry{Sequence}
		\addplot table[x=index,y=ntbag8, col sep= comma] {data/prefix_werk.csv};
		\addlegendentry{Bag}
		\addplot table[x=index,y=ntset8, col sep= comma] {data/prefix_werk.csv};
		\addlegendentry{Set}
		\legend{}
		\end{axis}
		\end{tikzpicture}
		\hfill
		\begin{tikzpicture}[scale=0.65]
		\begin{axis}[
		xlabel={Prefix Length},
		ylabel={Mean Absolute Error (days)},
		cycle list name=colorlist,
		]		
		\addplot table[x=index,y=ctBoW100, col sep= comma] {data/prefix_werk.csv};
		\addlegendentry{BoW}
		\addplot table[x=index,y=ctBoNG500, col sep= comma] {data/prefix_werk.csv};
		\addlegendentry{BoNG}
		\addplot table[x=index,y=ctPV100, col sep= comma]  {data/prefix_werk.csv};
		\addlegendentry{PV}
		\addplot table[x=index,y=ctLDA100, col sep= comma] {data/prefix_werk.csv};
		\addlegendentry{LDA}
		\addplot table[x=index,y=ct-0, col sep= comma] {data/prefix_werk.csv};
		\addlegendentry{LSTM}
		\addplot table[x=index,y=ctsequence8, col sep= comma] {data/prefix_werk.csv};
		\addlegendentry{Sequence}
		\addplot table[x=index,y=ctbag8, col sep= comma] {data/prefix_werk.csv};
		\addlegendentry{Bag}
		\addplot table[x=index,y=ctset8, col sep= comma] {data/prefix_werk.csv};
		\addlegendentry{Set}
		\legend{}
		\end{axis}
		\end{tikzpicture}
		\caption{\scriptsize BPIC2016 Customer Journey event log: next timestamp (left), cycle time (right).}
		\vspace{0.2cm}
	\end{subfigure}
	\begin{subfigure}{\textwidth}
		\centering
		\begin{tikzpicture}[scale=0.65]
		\begin{axis}[
		xlabel={Prefix Length},
		ylabel={Mean Absolute Error (days)},
		cycle list name=colorlist,
		]
		\addplot table[x=index,y=ntBoW500, col sep= comma] {data/prefix_admissions.csv};
		\addlegendentry{BoW}
		\addplot table[x=index,y=ntBoNG50, col sep= comma] {data/prefix_admissions.csv};
		\addlegendentry{BoNG}
		\addplot table[x=index,y=ntPV20, col sep= comma]  {data/prefix_admissions.csv};
		\addlegendentry{PV}
		\addplot table[x=index,y=ntLDA100, col sep= comma] {data/prefix_admissions.csv};
		\addlegendentry{LDA}
		\addplot table[x=index,y=nt-0, col sep= comma] {data/prefix_admissions.csv};
		\addlegendentry{LSTM}
		\addplot table[x=index,y=ntsequence8, col sep= comma] {data/prefix_admissions.csv};
		\addlegendentry{Sequence}
		\addplot table[x=index,y=ntbag8, col sep= comma] {data/prefix_admissions.csv};
		\addlegendentry{Bag}
		\addplot table[x=index,y=ntset8, col sep= comma] {data/prefix_admissions.csv};
		\addlegendentry{Set}
		\legend{}
		\end{axis}
		\end{tikzpicture}
		\hfill
		\begin{tikzpicture}[scale=0.65]
		\begin{axis}[
		xlabel={Prefix Length},
		ylabel={Mean Absolute Error (days)},
		cycle list name=colorlist,
		]
		\addplot table[x=index,y=ctBoW50, col sep= comma] {data/prefix_admissions.csv};
		\addlegendentry{BoW}
		\addplot table[x=index,y=ctBoNG100, col sep= comma] {data/prefix_admissions.csv};
		\addlegendentry{BoNG}
		\addplot table[x=index,y=ctPV100, col sep= comma]  {data/prefix_admissions.csv};
		\addlegendentry{PV}
		\addplot table[x=index,y=ctLDA10, col sep= comma] {data/prefix_admissions.csv};
		\addlegendentry{LDA}
		\addplot table[x=index,y=ct-0, col sep= comma] {data/prefix_admissions.csv};
		\addlegendentry{LSTM}
		\addplot table[x=index,y=ctsequence8, col sep= comma] {data/prefix_admissions.csv};
		\addlegendentry{Sequence}
		\addplot table[x=index,y=ctbag8, col sep= comma] {data/prefix_admissions.csv};
		\addlegendentry{Bag}
		\addplot table[x=index,y=ctset8, col sep= comma] {data/prefix_admissions.csv};
		\addlegendentry{Set}
		\legend{}
		\end{axis}
		\end{tikzpicture}
		\caption{\scriptsize MIMIC-III Hospital Admission event log: next timestamp (left), cycle time (right).}
	\end{subfigure}
	\caption{Prediction performance on selected metrics and logs, shown by length of trace prefix.}
	\label{fig:prefix}
\end{figure}

In addition, the prediction performance is evaluated per prefix length for each event log. Figure \ref{fig:prefix} shows the F$_1$ score and next timestamp MAE for every \clearpage prefix trace of length $1 \leq k \leq 8$ on a selection of prediction tasks. Note that the results on shorter traces are supported by a much larger set of traces due to prefix generation. For text-aware models, only the best encoding size is shown.

On the customer journey log, the performance of all models correlates positively with the available prefix length of the trace. All text-aware prediction models surpass the baseline approaches on very short prefix traces of length 3 or shorter, for next activity and outcome prediction: we hypothesize that the cause for this is a combination of higher availability of textual attributes in earlier events in the traces, and the high number of training samples of short lengths, which allow text models to generalize. The next timestamp and cycle time predictions show no difference between text-aware models and the LSTM baseline, although they systematically outperform transition system-based methods.

The hospital admission log is characterized by the alternation of admission and discharge events. Therefore, the prediction accuracy varies between odd and even prefix lengths. The text-aware prediction models generate slightly better predictions on admission events since only these contain the diagnosis as text attribute. Regarding the next timestamp prediction, higher errors after discharge events and low errors after admission events are observed.
This can be explained by the short hospital stays compared to longer time between two hospitalizations.

\section{Conclusion}\label{sec:conclusion}
The prediction of the future course of business processes is a major challenge in business process mining and process monitoring. When textual artifacts in a natural language like emails or documents hold critical information, purely control-flow-oriented approaches are limited in delivering accurate predictions.

To overcome these limitations, we propose a text-aware process predictive monitoring approach. Our model encodes process traces of historical process executions to sequences of meaningful event vectors using the control flow, time\-stamp, textual, and non-textual data attributes of the events.
Given an encoded prefix log of historical process executions, an LSTM neural network is trained to predict the activity and timestamp of the next event, and the outcome and cycle time of a running process instance. The proposed concept of text-aware predictive monitoring has been implemented and evaluated on real-world event data. We show that our approach is able to outperform state-of-the-art methods using insights from textual data.

The intersection between the fields of natural language processing and process mining is a promising avenue of research. Besides validating our approach on more datasets, future research also includes the design of a model able to learn text-aware trace and event embeddings, and the adoption of privacy-preserving analysis techniques able to avoid the disclosure of sensitive information contained in textual attributes.

\bibliographystyle{splncs04}
\bibliography{bibliography}

\end{document}